\documentclass{article} %
\usepackage{iclr2027_conference,times}

\usepackage{amsmath,amsfonts,bm}

\def\eqref#1{equation~\ref{#1}}

\def\1{\bm{1}}

\DeclareMathAlphabet{\mathsfit}{\encodingdefault}{\sfdefault}{m}{sl}
\SetMathAlphabet{\mathsfit}{bold}{\encodingdefault}{\sfdefault}{bx}{n}

\usepackage[english]{babel}
\usepackage{hyperref}
\hypersetup{colorlinks=true, linkcolor=[rgb]{0,0,0.55}, citecolor=[rgb]{0,0,0.55}, urlcolor=[rgb]{0,0,0.55}} %

\addto\extrasenglish{%
}
\usepackage{caption}
\usepackage{url}
\usepackage{xurl} %
\usepackage{graphicx}
\usepackage{tcolorbox}
\usepackage{xcolor}
\usepackage{amssymb}
\tcbuselibrary{skins} %
\tcbuselibrary{breakable}
\usepackage{enumitem}
\usepackage{colortbl}
\usepackage{array}
\usepackage{multirow}
\usepackage{longtable}
\usepackage{tabularx}

\usepackage{microtype}      %
\usepackage{booktabs}       %
\usepackage{wrapfig}        %
\usepackage[capitalise,nameinlink]{cleveref}

\definecolor{runexborder}{RGB}{0,32,96}
\newtcolorbox{runningexample}[1][Test]{
  enhanced, unbreakable,
  colback=white,
  colframe=runexborder,
  boxrule=1pt,
  sharp corners,
  fonttitle=\bfseries,
  coltitle=runexborder,
  colbacktitle=white,
  title=#1,
  attach boxed title to top left={xshift=10pt, yshift=-\tcboxedtitleheight/2},
  boxed title style={
    colback=white, colframe=white,
    boxrule=0pt, sharp corners,
    left=4pt, right=4pt, top=0pt, bottom=0pt,
  },
  top=10pt,
  left=10pt, right=4pt, bottom=4pt,
}
\newlist{researchq}{enumerate}{1}
\setlist[researchq]{
  label = RQ\arabic*.,
  ref   = RQ\arabic*,
  font  = \bfseries,
  widest = 9,
  align = left,
  leftmargin = *,
  labelsep = 0.5em,
}
\usepackage[table]{xcolor}
\usepackage{listings}
\usepackage{float}
\definecolor{sharedbg}{HTML}{C3EAE7}\definecolor{sharedfg}{HTML}{05706B}
\definecolor{astrabg}{HTML}{FADECF}\definecolor{astrafg}{HTML}{B1471A}
\definecolor{fablebg}{HTML}{E9DEF8}\definecolor{fablefg}{HTML}{7940AB}
\newcommand{\hlbox}[3]{{\setlength{\fboxsep}{0.5pt}\colorbox{#1}{\textcolor{#2}{\textbf{\strut #3}}}}}
\newcommand{\hlshared}[1]{\hlbox{sharedbg}{sharedfg}{#1}}
\newcommand{\hlastra}[1]{\hlbox{astrabg}{astrafg}{#1}}
\newcommand{\hlfable}[1]{\hlbox{fablebg}{fablefg}{#1}}
\lstdefinestyle{lstprompt}{basicstyle=\ttfamily\footnotesize,breaklines=true,breakatwhitespace=false,
  columns=fullflexible,keepspaces=true,frame=single,framerule=0.3pt,xleftmargin=0pt,
  breakindent=0pt,escapeinside={(*@}{@*)},literate={`}{\textasciigrave}1}

\title{Can Agents Design Libraries for Agents?}
\author{%
Gabriel Orlanski$^{1}$\thanks{Corresponding author: \texttt{gorlanski@cs.wisc.edu}}\hspace{1.5em}
Alex L. Zhang$^{2}$\hspace{1.5em}
Avi Trost$^{1}$\hspace{1.5em}
Vincent Sunn Chen$^{3}$ \\
\textbf{Frederic Sala$^{1,3}$\hspace{1.5em}
Aws Albarghouthi$^{1}$\hspace{1.5em}
Ludwig Schmidt$^{4}$} \\[0.5em]
$^{1}$University of Wisconsin--Madison\hspace{1.5em}
$^{2}$Massachusetts Institute of Technology \\
$^{3}$Snorkel AI\hspace{1.5em}
$^{4}$Stanford University
}

\newcommand{\ldb}{LibraryDesignBench}
\newcommand{\phaseone}{Design Phase}
\newcommand{\phasetwo}{Evaluation Phase}
\newcommand{\clirs}{\texttt{clirs}}

\newcommand{\agent}{\pi_\theta}
\newcommand{\implementer}[1][u]{\pi_{#1}}
\newcommand{\implementers}{U}

\newcommand{\library}{L}
\newcommand{\ceilingLib}{L^{\star}}

\newcommand{\instruction}{I}
\newcommand{\problem}[1]{x_{#1}}
\newcommand{\evalproblems}{\mathcal{P}}
\newcommand{\solution}[1]{y_{#1}}

\newcommand{\numruns}{K} %
\newcommand{\ceilingRef}[1]{#1^{*}}

\newcommand{\ceilingSol}[1][]{\ceilingRef{\solution{#1}}}
\newcommand{\metricset}{M}
\newcommand{\passrate}{q}
\newcommand{\libscore}[1][\library{}]{\operatorname{score}(#1)}
\newcommand{\simplicity}{\rho}

\newcommand{\numtasks}{\mathcal{T}}
\newcommand{\benchscore}{\widehat{S}}
\newcommand{\clusterse}[1]{%
    \widehat{\operatorname{SE}}_{\mathrm{cl}}\!\left(#1\right)%
}
\newcommand{\rerunse}[1]{%
    \widehat{\operatorname{SE}}_{\mathrm{run}}\!\left(#1\right)%
}

\newcommand{\repolink}{https://github.com/SprocketLab/librarydesignbench}

\newcommand{\dsproworsenolib}{9.2} %
\newcommand{\existinglibrarymeanscore}{46.6} %
\newcommand{\hardtousepctsampled}{64\%} %
\newcommand{\haskellnolibbetterpct}{70} %
\newcommand{\lowhighreasoningincreasepct}{63\%} %
\newcommand{\lubmodels}{8} %
\newcommand{\neuraleseauthorcost}{\$8.24} %
\newcommand{\neuralesecostchangepm}{\ensuremath{-\$0.01 \pm \$0.01}} %
\newcommand{\neuraleseimprovement}{2.3} %
\newcommand{\neuraleseincumbentrecall}{13.4} %
\newcommand{\neuralesepassrateloss}{0.9} %
\newcommand{\neuralesepctsimplicityimprove}{6.8} %
\newcommand{\neuralesescore}{46.4} %
\newcommand{\neuralesesimplicitygain}{2.5} %
\newcommand{\neuralesestandardauthorcost}{\$4.51} %
\newcommand{\neuralesestandardscore}{44.1} %
\newcommand{\nolibrarypassratediff}{0.2} %
\newcommand{\numcasesnolibbetter}{3} %
\newcommand{\numlubevaluated}{8} %
\newcommand{\passratemax}{86.6} %
\newcommand{\passratemin}{84.1} %
\newcommand{\prescriptionpctincreass}{23\%} %
\newcommand{\productionlibrarycostincreasepm}{\ensuremath{\$0.14 \pm \$0.02}} %
\newcommand{\productionlibraryimprovepm}{\ensuremath{12.3 \pm 0.4}} %
\newcommand{\sotalubmodel}{Opus 5.5} %
\newcommand{\sotalubscore}{66.9} %
\newcommand{\sotamodel}{Opus 5.5} %
\newcommand{\sotaoverproductionlibrary}{2.3} %
\newcommand{\sotascore}{48.9} %
\newcommand{\sotascoreimprovepct}{4.9} %
\newcommand{\sotaworsenolibmeanpp}{3.2} %
\newcommand{\standardincumbentrecall}{19.4} %
\newcommand{\taxauthors}{6} %
\newcommand{\taxcellsperlibrary}{3} %
\newcommand{\taxexcessatceilingpct}{82\%} %
\newcommand{\taxexcesscases}{810} %
\newcommand{\taxexcesscoveragepct}{14\%} %
\newcommand{\taxexcessergonomicspct}{64\%} %
\newcommand{\taxfailedcases}{810} %

\iclrfinalcopy %
\begin{document}

\maketitle
\lhead{}  %

\begin{abstract}
Agents increasingly build on code written by other agents, and they reimplement rather than reuse, growing the codebases later agents must work in.
To measure how well agents design libraries for other agents, we introduce LibraryDesignBench, a two-phase benchmark in which an agent implements a full-featured library from a specification that defines required capabilities and potential use cases without prescribing the design.
We evaluate the library through the correctness and simplicity of programs written by three user agents from different model families.
The benchmark spans 242 expert-validated programming problems across 15 library-design tasks in four languages. On eleven of the fifteen tasks, agent designers reproduce the abstractions of the human-written production library. Downstream agents adopt agent- and human-written libraries alike but underuse them, reimplementing capabilities the library already provides.
Our failure analysis finds that downstream agents write extra code mainly because agent-written libraries are rigid or hard to use, not because capabilities are missing.
We also experiment with giving designers more prescriptive, agent-first guidance and having them test their library with subagents; this improves downstream scores and yields simpler programs.
LibraryDesignBench provides both a testbed for evaluating library-design practices for agent users and an initial design baseline that improves downstream reuse.
\end{abstract}
\section{Introduction}
Software engineering as a discipline would not exist without skilled engineers designing libraries, frameworks, SDKs, etc., with \textit{opinionated design decisions} that make future work easier. Yet, we are fast approaching an inflection point where agents will work more with code designed by another agent than by a human engineer. This raises a fundamental question. \textit{Can agents design libraries that other agents can leverage?} Poorly designed libraries will hamper future agents' performance in both correctness and code volume -- requiring more human effort and intervention to repair. Answering it demands evaluating the library through downstream agent use, not correctness tests alone.

\begin{figure}[h]
    \centering
    \includegraphics[width=1\linewidth]{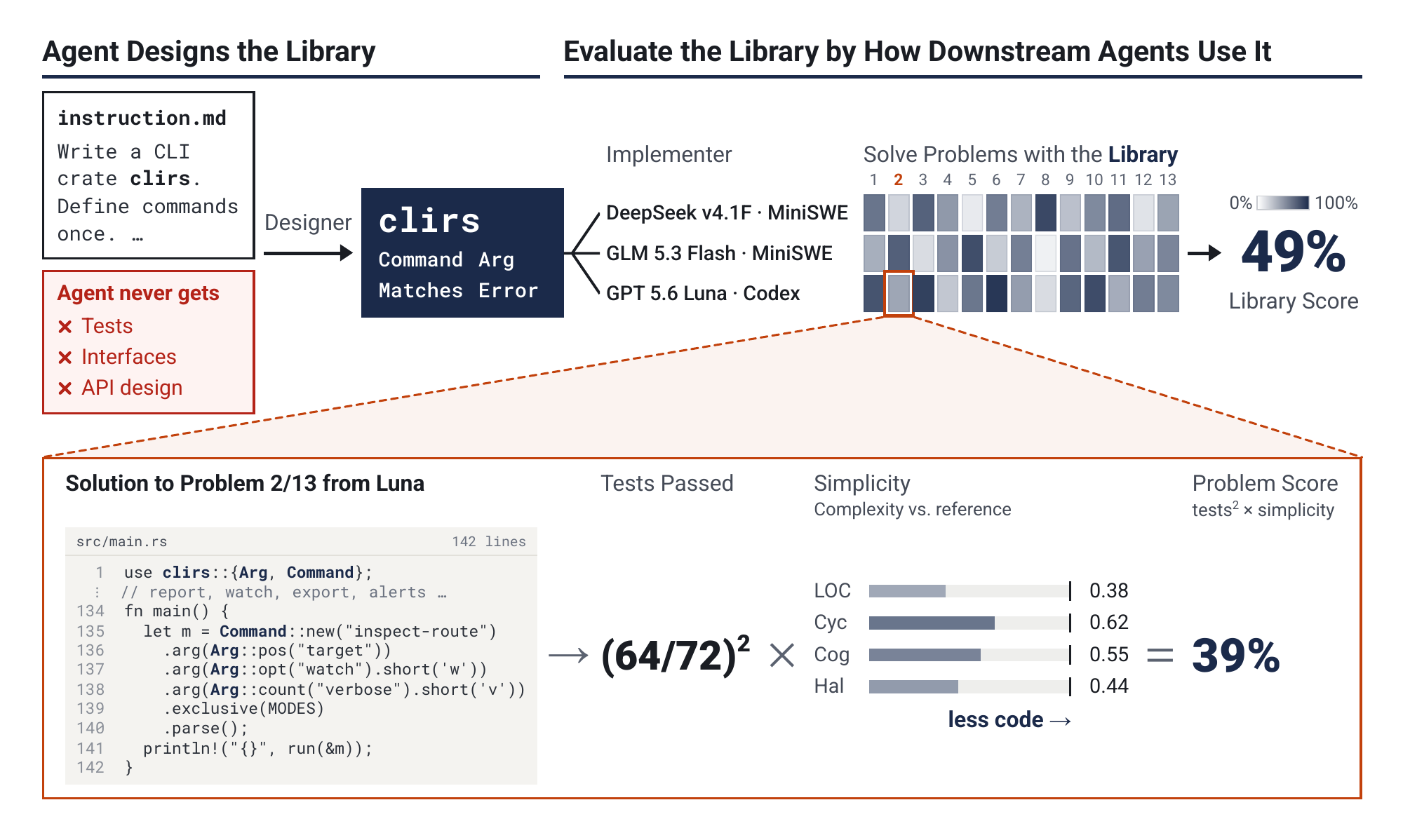}
    \caption{\textbf{LibraryDesignBench's two-phase setup evaluates the library through real usage.} The agent under evaluation designs the library from a non-prescriptive specification. Three downstream agents then solve problems with it. The score reflects how correct and simple their programs are.}
    \label{fig:hero}
\end{figure}
The core roadblock is that grading a generated library is nontrivial. Tests only check that the library is correct, not whether it helps the agents that use it. Grading the interface against a specification requires signatures or constructs, which prescribe the very design we want to measure~\citep{nl2repo,commit0,projecteval,icae,refactorbench}. Agentic validation~\citep{seniorswe} and static metrics score the implementation, not its usability. Human review measures what humans prefer, not what agents need. \textit{The only faithful way to evaluate a library written for agents is to observe agents use it.}

We therefore propose \textbf{LibraryDesignBench}, a two-phase evaluation that assesses an agent's ability to design a library by measuring how downstream agents use it to write better programs.
The benchmark comprises fifteen library-design problems and 242 expert-validated downstream programming problems across four languages.
In the \textbf{\phaseone{}}, the agent under evaluation implements a full-featured library from a specification that defines required capabilities while leaving interfaces and abstractions open.
In the \textbf{\phasetwo{}}, multiple different, less capable user agents use that library to implement downstream programs.
We evaluate the library by the correctness and simplicity of these programs, using reference solutions built with real production libraries. We define simplicity using capped reference-to-program ratios averaged over four static size and complexity measures.

\sotamodel{} scores highest (\sotascore{}), \sotaoverproductionlibrary{} points above the production library. In eleven of fifteen tasks, designers reproduce production-library abstractions. Our audit classifies \hardtousepctsampled{} of sampled excess-code cases under rigid or hard-to-use interfaces, and only \taxexcesscoveragepct{} under missing capabilities.

A library's score also depends on how well implementers use it. Even with the production library, implementers reach a simplicity of only 61.5, so part of the gap to the reference comes from the implementer, not the design. To measure this separately, we fix the library to the production library and evaluate \numlubevaluated{} models as implementers, which we call \textbf{LibraryUseBench}. \sotalubmodel{} scores best at \sotalubscore{}. For GPT-5.6 Luna, higher reasoning effort mainly improves correctness, a more prescriptive prompt mainly improves library use, and its solutions remain far longer than the reference.

Because agents currently reproduce production-library abstractions, we next test whether more prescriptive agent-first guidance helps. We have GPT-6 Astra sketch consumer programs first, ship them as runnable usage examples, and test its design with subagents. This raises the score by \neuraleseimprovement{} points, mainly from \neuralesepctsimplicityimprove{}\% higher simplicity, yet it still falls just short of the production library. \textit{Designing libraries for agents differs from designing them for humans, and remains an open problem.}

Our contributions are:

\begin{itemize}[ widest = 9,
  align = left,
  leftmargin = *,
  labelsep = 0.5em]
    \item \textbf{LibraryDesignBench.} We introduce \ldb{} which evaluates agent-designed libraries only by observing how well downstream agents can leverage them. (\autoref{sec:ldb})
    \item \textbf{How agents design and use libraries, and why they fail.} Agent-designed libraries reproduce production-library abstractions (\autoref{sec:overall-results}), and our audit attributes most excess-code failures to rigid or hard-to-use interfaces, not missing capabilities (\autoref{sec:failure-taxonomy}). Even with a production library, agents exploit it only when pushed and still write far more code than an expert (\autoref{sec:lib-usage}).
    \item \textbf{Prompting Interventions.} More prescriptive agent-first guidance, combining consumer-first API sketches, runnable usage examples, and testing with subagents, reduces exported-name overlap with production libraries, improves downstream scores, and yields simpler programs (\autoref{sec:agent-preferences}).\footnote{Code and data: \url{\repolink}}\label{fn:repo}
\end{itemize}

\section{LibraryDesignBench}\label{sec:ldb}

The core goal of \ldb{} is to measure the quality of a library designed by an agent \textit{only} by observing how downstream agents utilize it. A single task consists of two phases:
\begin{itemize}
    \item \textbf{\phaseone{}}: tasks the \textit{agent under evaluation}, $\agent{}$, with building the library $\library{}$.
    \item \textbf{\phasetwo{}}: downstream implementers, $\implementer{}$, solve tasks using $\library{}$.
\end{itemize}

To score $\library$, we consider both \textit{correctness} through the task's test suite and \textit{simplicity} compared to reference solutions written idiomatically with a real production library. 

\begin{runningexample}[Running problem: CLI crate for Rust]
\label{sec:running-problem}
\textbf{\phaseone{}:} Design \clirs{}, a Rust CLI-parsing crate.  The spec
names commands, flags, options, typed values, arity, cross-argument relations,
and generated help.  The environment is offline with no argument-parsing
library, so the agent builds the parser from scratch.

\textbf{\phasetwo{}:} A fresh agent builds one of 13 CLI tools with the authored crate.  \texttt{site-plan}, for instance, needs two subcommands sharing a group of lifetime options of which exactly one must be given, a repeatable \texttt{-{}-resource SRC DEST} with fixed two-value arity, four equivalent option spellings (\texttt{-ndocs}, \texttt{-n=docs}, \texttt{-{}-site-name=docs}, \texttt{-n~docs}), and grouped 72-column help. Other tools stress nested command trees with aliases, inherited globals, reusable argument groups, and parser-consistent introspection.

\textbf{Production library:}
\href{https://docs.rs/clap/latest/clap/}{clap}.  Behavioral tests establish
correctness. Program size relative to the reference clap solution measures
library value.
\end{runningexample}

\subsection{Evaluating Libraries Through Real Observation}\label{sec:two-phase}

\ldb{} measures how well an agent can create a library, $\library{}$, that helps the future agents who use it. Direct test suites, or even agentic verifiers, can only measure \textit{whether this library is correct}. They cannot measure how it will impact agents trying to use it to solve \textit{real} tasks.

\paragraph{\phaseone{}.} The agent under evaluation, $\agent$, implements $\library{}$ given an instruction, $\instruction{}$. The instruction leaves interfaces and abstractions open, forcing $\agent$ to reason through which abstractions are needed and which are not. It is provided a list of functionality it must support and two to three example usages drawn from the task's own evaluation problems. These ``visible'' problems give the agent the grounding it needs for understanding how its library will be used, similar to how real engineers express a library spec. The visible problems remain in the scored evaluation set, analogous to visible test cases. Beyond the named capabilities, the general instructions require functionality users would reasonably expect of the library (\autoref{sec:packaging}). We also instruct agents that the primary users of $\library{}$ will be coding agents and that senior engineers will review their work. Finally, agents must package their library with a language-specific package manager so that it installs.

\paragraph{\phasetwo{}.} Each task contains a set of problems, $\evalproblems$, that each implementer $\implementer{} \in \implementers{}$ will solve using $\library$.
Problems must be solvable with or without a library. Implementers are explicitly instructed to make their solutions ``thin adapters'' over the libraries and to ensure they read the documentation (\autoref{lst:main-prompt}). We want to elicit the library's ability to be exploited to write as little code as possible, not the implementer's ability to recognize when a library helps. 

\subsection{Scoring the Quality of a Library}\label{sec:score}
A library is only valuable if the underlying implementations are \textit{correct} and it enables writing \textit{simpler} programs. A library whose programs are shorter but incorrect, or correct but no shorter than without it, provides little value. Thus, we \textit{score} a $\library{}$ according to the product of correctness and simplicity:
\begin{equation}\label{eq:score}
    \begin{aligned}
        \libscore
        &= \frac{1}{|\evalproblems|\cdot|\implementers|}
           \sum_{u\in\implementers}
           \sum_{\problem{i}\in\evalproblems}
           \passrate_i(\solution{i})^2
      \cdot\simplicity_i(\solution{i}),
        &\qquad \solution{i}\sim\implementer[u](\problem{i}\mid\library).
    \end{aligned}
\end{equation}
Here $\passrate_i(\solution{i})\in[0,1]$ is the fraction of tests passed and $\simplicity_i(\solution{i})$ measures static simplicity relative to the problem's reference solution. Each implementer produces one solution per problem, and we report scores multiplied by 100. We square the test-pass fraction to prioritize correctness while retaining graded credit for partially correct programs, so passing 80\% of tests at the reference's size ($0.64$) scores below passing every test at $1.5\times$ its size ($0.67$). If $\library{}$ cannot be installed, its score is always 0 across all problems. \autoref{sec:aggregation} derives the aggregation, standard errors, and confidence intervals.

\paragraph{Simplicity.}
A library should reduce the code needed to solve a task, first and foremost. Let $\ceilingSol[i]$ be the fixed optimized reference solution for problem $i$. We define
\begin{equation}\label{eq:simplicity}
    \simplicity_i(\solution{i})
    = \frac{1}{|\metricset|}
      \sum_{m\in\metricset}
      \min\!\left\{
          \frac{m(\ceilingSol[i])}{m(\solution{i})},\,1
      \right\}.
\end{equation}
We utilize a set of static metrics, $\metricset$, which reduces dependence on any single static metric (e.g., a parser that accepts every option spelling removes branches, not just lines). Averaging over $\metricset$ keeps $\simplicity_i$ on the same scale as a single metric ratio, so a solution that matches its reference on every metric scores $1$ and one twice its size on every metric scores $0.5$. Capping each ratio at one bounds the contribution of programs smaller than the reference. If a metric is zero for the solution, its ratio is set to 1. If no solution is produced, or the solution has syntax errors, its simplicity is zero. 

$\metricset$ consists of static counts that quantify residual code without grading conformity to an interface:
\[
    \begin{aligned}
        \metricset=\{\text{Cyclomatic Complexity},
                      \text{Cognitive Complexity},
                    \text{Halstead Volume},
                      \text{Source Lines of Code}\},
    \end{aligned}
\]
with definitions and language-specific counting rules detailed in \autoref{appendix:staticmeasures} for Cyclomatic Complexity~\citep{mcabecyc}, Cognitive Complexity~\citep{cogcomplex}, and
Halstead Volume~\citep{Halstead1977ElementsOS}. We measure source lines of code after applying a language-standard formatter, excluding comments and blank lines. Metrics are computed over eligible source files using the counting and aggregation rules in \autoref{appendix:staticmeasures}. Cognitive complexity targets human comprehension, but it captures a different kind of complexity than cyclomatic, and agents spend more tokens and revisit more files on code that scores high on it and violates more static-analysis rules~\citep{Trivedi2026DoesCC}.

\paragraph{Reference solutions.} The reference $\ceilingSol[i]$ is a fixed program per problem written using the mature production library $\ceilingLib{}$ and optimized for idiomatic use of its abstractions, rather than code golfing. Library experts developed and optimized the reference programs with agent assistance. We apply the same formatting and measurement procedures to reference and generated programs. For example, the \clirs{} references use \texttt{clap}'s declarative \texttt{derive} pattern, not its builder syntax. 

\paragraph{Implementer configurations.} Each $u\in\implementers$ identifies a complete implementer configuration, including its model, harness, and inference settings. We use the same fixed set across all generated libraries and comparison conditions. \autoref{eq:score} intentionally weights every implementer equally.

\subsection{Benchmark Aggregation and Reporting}\label{sec:aggregation}
To evaluate the designer rather than a single artifact, we independently repeat the design phase $\numruns=3$ times for each of the $\numtasks$ benchmark tasks. We evaluate each resulting library on its task's problems using the same fixed implementer set, with fresh downstream executions. We average all generations rather than selecting the best, reducing the influence of an unusually successful or unsuccessful run. The production-library and no-library settings have no design phase. For them, we repeat the evaluation phase $\numruns=3$ times, and $k$ indexes these repetitions.

Let $\library_{p,k}$ be the library generated for task $p$ on run $k$, and let
\begin{equation}\label{eq:library-run-score}
    S_{p,k}
    = \libscore[\library_{p,k}],
    \qquad
    \bar S_p
    = \frac{1}{\numruns}
      \sum_{k=1}^{\numruns}S_{p,k}.
\end{equation}
Here, $S_{p,k}$ retains the average over problems and implementers defined in \autoref{eq:score}. We then give each library-design task equal weight:
\begin{equation}\label{eq:benchmark-score}
    \benchscore
    = \frac{1}{\numtasks}
      \sum_{p=1}^{\numtasks}\bar S_p
    = \frac{1}{\numtasks\numruns}
      \sum_{p=1}^{\numtasks}
      \sum_{k=1}^{\numruns}S_{p,k}.
\end{equation}
This prevents tasks with more downstream problems from receiving greater benchmark weight.

\paragraph{Rerun standard errors.} To quantify how much the reported score would fluctuate, given the stochastic nature of agentic evaluation, we formulate a standard error for agent-to-agent evaluations. Each independently generated library together
with its complete downstream evaluation constitutes one observation. 

These observations need not be identically distributed across tasks as different tasks have disparate expected scores and execution variances. We therefore estimate variability \textit{within} each task, treating tasks as fixed strata rather than measuring deviations around a single overall mean. For each task, the sample variance across library runs is
\begin{equation}\label{eq:within-task-variance}
    s_p^2
    =
    \frac{1}{\numruns-1}
    \sum_{k=1}^{\numruns}
    (S_{p,k}-\bar S_p)^2.
\end{equation}
Assuming independent, identically distributed repetitions within each task and independence across tasks, the estimated standard error of the benchmark mean is
\begin{equation}\label{eq:rerun-se}
    \rerunse{\benchscore}
    =
    \frac{1}{\numtasks}
    \sqrt{
        \sum_{p=1}^{\numtasks}
        \frac{s_p^2}{\numruns}
    }.
\end{equation}
The squared expression is an unbiased estimator of the variance of \autoref{eq:benchmark-score}, provided the run scores have finite variance. Each library's deviation is measured relative to its own task mean, so stable differences between tasks do not contribute.

We retain dependence within a library evaluation by computing $S_{p,k}$ before estimating its variance. For example, an architectural defect can hurt several problems or implementers simultaneously. These shared effects contribute to the variance of the complete library score, so we do not treat the downstream programs as independent library observations. This follows the principle of retaining related evaluations together when estimating uncertainty~\citep{miller2024adding}.

\paragraph{Confidence intervals.} We report approximate $95\%$ confidence intervals for the designer's expected score under repeated execution of this fixed evaluation:
\[
    \mu
    =
    \frac{1}{\numtasks}
    \sum_{p=1}^{\numtasks}\mathbb{E}[S_{p,1}].
\]
Because the task-specific variances are estimated from a small number of runs, we use a Student-$t$ interval with Welch--Satterthwaite effective degrees of freedom:
\begin{equation}\label{eq:effective-df}
    \widehat\nu
    =
    \frac{
        \left(
            \sum_{p=1}^{\numtasks}s_p^2/\numruns
        \right)^2
    }{
        \sum_{p=1}^{\numtasks}
        \frac{
            (s_p^2/\numruns)^2
        }{\numruns-1}
    }.
\end{equation}
The reported interval is
\begin{equation}\label{eq:rerun-ci}
    \benchscore
    \;\pm\;
    t_{\widehat\nu,\,0.975}
    \,\rerunse{\benchscore},
\end{equation}
where $t_{\nu,\,0.975}$ is the $97.5$th percentile of a Student-$t$ distribution with $\nu$ degrees of freedom. The interval is a model-based approximation motivated by approximately normal within-task run-score distributions. With only three generations per task, nominal coverage is not guaranteed. The interval reflects stochastic variation in both phases on this fixed benchmark.

\paragraph{Descriptive standard errors.} The $\pm$ values reported for pass rate, simplicity, cost, and tokens are standard errors of the mean clustered by task, $\clusterse{\cdot}$. They describe variation across problems and runs rather than the rerun uncertainty of the score. A reported difference between two conditions' descriptive means (e.g., the change in cost per problem) combines the two conditions' clustered standard errors in quadrature. Score differences between conditions instead combine the two conditions' rerun standard errors, $\rerunse{\cdot}$, in quadrature.

\subsection{Benchmark Construction}\label{sec:construction}

We now detail the construction of \ldb{}, which yielded fifteen tasks across four programming languages. We selected libraries to use as tasks based on their age, complexity, and the number of interface decisions a designer must make. The \phasetwo{} problem desiderata are:
\begin{enumerate}
\item \textbf{Realistic task.} A problem reflects a realistic use case for the library.
\item \textbf{Library Headroom.} A problem is valuable to \ldb{} if the library reduces significant amounts of code through composition and interaction of features.
\item \textbf{Solvable Without a Library.} A problem written such that only one library could reasonably solve it is not a fair problem for \ldb{}. Thus, every problem must be solvable without any library available.
\end{enumerate}

Each problem is built through a multi-stage agentic pipeline that isolates the library's functionality, seeded by real usages of the library from permissively licensed repositories. First, an extraction agent reduces the seed to a minimal program that exercises the library, removing application-specific logic. Next, a rewriting agent produces a library-free equivalent and a test suite on which both implementations must agree. We then verify that every test is solvable from the task instructions and workspace alone. Finally, library experts review each problem, rewrite its instructions, and strengthen its tests against solutions that omit required behavior. A second expert audits each review.

\section{Evaluating Frontier Models on LibraryDesignBench}
Each designer runs in mini-SWE-agent unless noted, with $\numruns{}=3$ libraries per task and 3 implementers per library (2,178 evaluated problems per designer). We answer four research questions:
\begin{researchq}
  \item\label{rq:agent-design} \textbf{Can agents design libraries that improve other agents?} Yes. The strongest designer exceeds the production-library baseline by \sotaoverproductionlibrary{} points (\sotascoreimprovepct{}\% relative), while designers reproduce production-library abstractions on eleven of fifteen tasks. (\autoref{sec:overall-results})
  \item\label{rq:failure-taxonomy} \textbf{Why do agents struggle with agent-designed libraries?} In sampled partially passing solutions, our audit classifies \hardtousepctsampled{} of excess-code cases under rigid or hard-to-use interfaces, not missing capabilities. (\autoref{sec:failure-taxonomy})
  \item\label{rq:library-use} \textbf{How can implementers better leverage libraries?} More prescriptive prompts and higher reasoning effort raise the score by \prescriptionpctincreass{} and \lowhighreasoningincreasepct{} relative. Prescription drives library use while effort drives correctness, and solutions stay far longer than the reference. (\autoref{sec:lib-usage})
  \item\label{rq:prescription} \textbf{Can agentic design patterns improve implementer performance?} Yes, modestly. More prescriptive agent-first guidance raises the score by \neuraleseimprovement{} points. (\autoref{sec:agent-preferences})
\end{researchq}

\begin{table}[t]
     \centering
     \small
     \caption{\textbf{Overall results for each library setup by designer.} All results average the same implementer set. Designers use mini-SWE-agent~\citep{mswea} unless another harness is named in parentheses, where CC is Claude Code. Production and No library are settings in which the implementer is given a human-written library or no library at all (\autoref{lst:prompt-nolib}), respectively. ``Library \$'' is the average cost, in USD, to generate a single library, while ``Problem \$'' is the average implementer cost per evaluation problem. ``\% Pass'' is the mean share of tests passed, not of fully solved problems. Scores show 95\% CIs. Other values show clustered standard errors (\autoref{sec:aggregation}). \textbf{Bold} marks the best per column.}
\label{tab:overall-results}
         \begin{tabular}{l|rrrrr}
    \toprule
    Model (Harness) & Score ($\uparrow$) & \% Pass ($\uparrow$) & Simplicity ($\uparrow$) & Library \$ ($\downarrow$) & Problem \$ ($\downarrow$) \\
    \midrule
    DeepSeek V4 Pro & 31.2 {\scriptsize[30.4, 32.1]} & 84.9 {\scriptsize$\pm$1.7} & 42.0 {\scriptsize$\pm$1.1} & \$0.31 {\scriptsize$\pm$0.02} & \$0.175 {\scriptsize$\pm$0.008} \\ %
    Fable 5.1 & 47.5 {\scriptsize[46.4, 48.5]} & 86.1 {\scriptsize$\pm$1.6} & 62.7 {\scriptsize$\pm$1.8} & \$14.81 {\scriptsize$\pm$1.37} & \$0.198 {\scriptsize$\pm$0.013} \\ %
    Fable 5.1 (CC) & 39.9 {\scriptsize[37.4, 42.4]} & 84.9 {\scriptsize$\pm$1.6} & 58.3 {\scriptsize$\pm$2.5} & \$14.39 {\scriptsize$\pm$2.23} & \$0.211 {\scriptsize$\pm$0.014} \\ %
    GLM 5.3 & 41.8 {\scriptsize[40.8, 42.9]} & 84.1 {\scriptsize$\pm$1.7} & 57.1 {\scriptsize$\pm$1.7} & \$21.06 {\scriptsize$\pm$1.89} & \$0.234 {\scriptsize$\pm$0.013} \\ %
    GPT-5.6 Sol (Codex) & 39.5 {\scriptsize[38.4, 40.5]} & 84.1 {\scriptsize$\pm$2.1} & 52.9 {\scriptsize$\pm$1.2} & \$2.14 {\scriptsize$\pm$0.20} & \$0.199 {\scriptsize$\pm$0.011} \\ %
    GPT-6 Astra & 45.1 {\scriptsize[44.3, 45.8]} & 85.7 {\scriptsize$\pm$1.7} & 58.7 {\scriptsize$\pm$1.5} & \$3.63 {\scriptsize$\pm$0.24} & \$0.155 {\scriptsize$\pm$0.008} \\ %
    GPT-6 Astra (Codex) & 44.1 {\scriptsize[43.4, 44.9]} & 85.4 {\scriptsize$\pm$1.9} & 58.5 {\scriptsize$\pm$1.6} & \$4.51 {\scriptsize$\pm$0.40} & \$0.194 {\scriptsize$\pm$0.012} \\ %
    GPT-6 Sol & 38.5 {\scriptsize[37.8, 39.2]} & 85.2 {\scriptsize$\pm$1.9} & 51.6 {\scriptsize$\pm$1.3} & \textbf{\$0.30} {\scriptsize$\pm$0.02} & \$0.150 {\scriptsize$\pm$0.007} \\ %
    Grok 4.6 & 39.7 {\scriptsize[38.6, 40.7]} & 84.6 {\scriptsize$\pm$1.6} & 54.1 {\scriptsize$\pm$1.6} & \$2.63 {\scriptsize$\pm$0.23} & \$0.225 {\scriptsize$\pm$0.012} \\ %
    Kimi K3 & 44.0 {\scriptsize[43.0, 45.0]} & 86.0 {\scriptsize$\pm$1.5} & 58.6 {\scriptsize$\pm$1.7} & \$8.70 {\scriptsize$\pm$1.47} & \$0.176 {\scriptsize$\pm$0.010} \\ %
    Opus 5.5 & \textbf{48.9} {\scriptsize[48.0, 49.9]} & \textbf{86.6} {\scriptsize$\pm$1.4} & \textbf{64.5} {\scriptsize$\pm$1.8} & \$9.62 {\scriptsize$\pm$1.03} & \$0.174 {\scriptsize$\pm$0.011} \\ %
    \midrule
    Production Library & 46.6 {\scriptsize[46.1, 47.2]} & 85.4 {\scriptsize$\pm$1.0} & 61.5 {\scriptsize$\pm$0.9} & -- & \$0.240 {\scriptsize$\pm$0.020} \\ %
    No Library & 34.4 {\scriptsize[33.9, 34.9]} & 86.4 {\scriptsize$\pm$1.0} & 46.3 {\scriptsize$\pm$1.1} & -- & \textbf{\$0.098} {\scriptsize$\pm$0.006} \\ %
    \bottomrule
    \end{tabular}

\end{table}

\paragraph{Setup.} Agents have no internet access, 4 hours to design, and 1 hour and \$2.50 per problem. Unfinished solutions are scored as is, which affects 1.7\% of trials (\autoref{appendix:timeouts}). Details are in \autoref{appendix:setup}.

\subsection{Agent Design Quality}\label{sec:overall-results}

\begin{figure*}[h]
  \centering
  \includegraphics[width=1\linewidth]{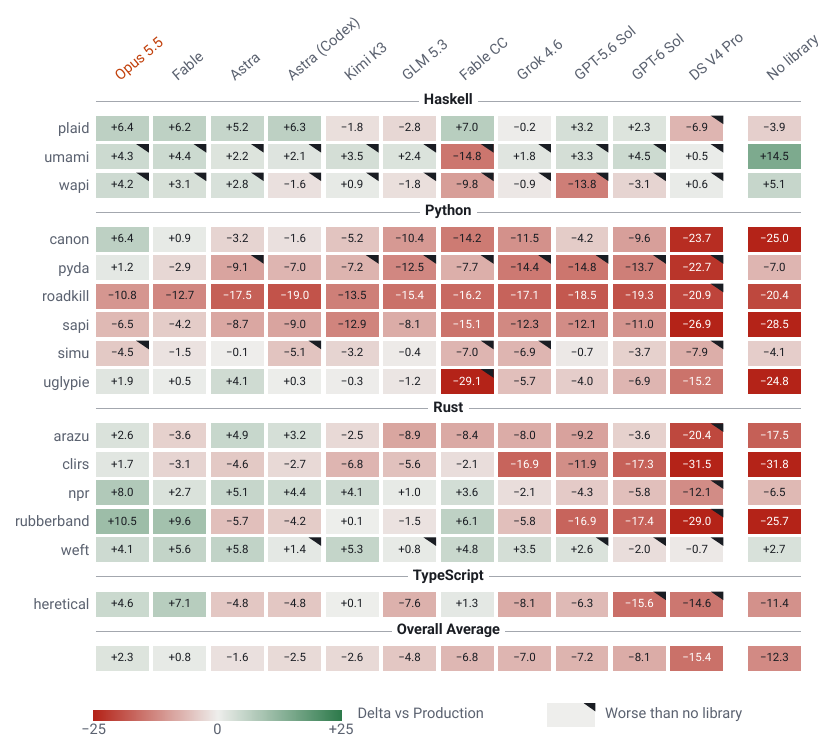}
  \caption{\textbf{Agent-written libraries outperform production libraries on a subset of tasks.} Difference in score per task compared with that of the production library. The rightmost column is the no-library setup. Outlined cells score below no library.}
  \label{fig:tasks}
\end{figure*}

\autoref{tab:overall-results} highlights our overall results. Downstream correctness does not separate designers, as every setup passes \passratemin\%--\passratemax\% of tests. The no-library condition reaches 86.4\%, within \nolibrarypassratediff{} percentage points of the highest mean test-pass rate. Score differences primarily reflect how simple downstream agents' solutions are. Production libraries add $\productionlibraryimprovepm$ points over no library at $\productionlibrarycostincreasepm$ more per problem. On the other end, DeepSeek V4 Pro's library scores \dsproworsenolib{}\% below no library. Harm is most common in Haskell, where agent-written libraries score below no library in about \haskellnolibbetterpct{}\% of the 33 (designer, Haskell task) pairs. On the \numcasesnolibbetter{} tasks where no library beats production, \sotamodel{}'s libraries trail no library by \sotaworsenolibmeanpp{} points (\autoref{fig:tasks}). The harness also matters. Fable 5.1 scores 47.5 in mini-SWE-agent but 39.9 in Claude Code. We therefore report harness variants separately.

\begin{table}[t]
     \centering
     \small
     \setlength{\tabcolsep}{4pt}
    \caption{\textbf{Results per implementer across library setups.} Only mini-SWE-agent designers are shown. ``Problem tokens'' is the mean number of tokens per problem.}
\label{tab:arm-ladder}
        \begin{tabular}{l|rrrr}
    \toprule
    Setup & Score (95\% CI, $\uparrow$) & \% Pass ($\uparrow$) & Simplicity ($\uparrow$) & Problem tokens ($\downarrow$) \\
    \midrule
    \rowcolor[gray]{0.93} \multicolumn{5}{c}{\emph{Implementer:} \textbf{DeepSeek V4.1 Flash}} \\
    \quad Astra & 46.9 {\scriptsize[45.8, 47.9]} & 87.4 {\scriptsize$\pm$1.7} & 58.5 {\scriptsize$\pm$1.4} & 5.80 {\scriptsize$\pm$0.34}M \\
    \quad GPT-6 Sol & 41.3 {\scriptsize[40.2, 42.5]} & 87.0 {\scriptsize$\pm$1.8} & 51.6 {\scriptsize$\pm$1.2} & 4.86 {\scriptsize$\pm$0.28}M \\
    \quad Fable & 50.2 {\scriptsize[48.9, 51.5]} & 87.8 {\scriptsize$\pm$1.7} & 63.2 {\scriptsize$\pm$1.8} & 6.35 {\scriptsize$\pm$0.35}M \\
    \quad Opus 5.5 & \textbf{51.6} {\scriptsize[50.4, 52.8]} & 88.4 {\scriptsize$\pm$1.5} & \textbf{64.3} {\scriptsize$\pm$1.8} & 6.47 {\scriptsize$\pm$0.36}M \\
    \quad GLM 5.3 & 45.1 {\scriptsize[43.8, 46.4]} & 86.2 {\scriptsize$\pm$1.8} & 57.3 {\scriptsize$\pm$1.6} & 8.43 {\scriptsize$\pm$0.46}M \\
    \quad Grok & 42.1 {\scriptsize[40.7, 43.4]} & 86.6 {\scriptsize$\pm$1.7} & 53.8 {\scriptsize$\pm$1.4} & 7.07 {\scriptsize$\pm$0.42}M \\
    \quad DeepSeek V4 Pro & 33.3 {\scriptsize[32.4, 34.3]} & 87.5 {\scriptsize$\pm$1.8} & 41.3 {\scriptsize$\pm$1.3} & 5.92 {\scriptsize$\pm$0.29}M \\
    \quad Kimi K3 & 46.4 {\scriptsize[45.2, 47.7]} & 87.8 {\scriptsize$\pm$1.5} & 57.9 {\scriptsize$\pm$1.6} & 6.23 {\scriptsize$\pm$0.35}M \\
    \cmidrule(lr){1-5}
    \quad Production library & 49.3 {\scriptsize[48.5, 50.2]} & 86.2 {\scriptsize$\pm$3.8} & 62.0 {\scriptsize$\pm$2.4} & 6.49 {\scriptsize$\pm$0.97}M \\
    \quad No library & 37.0 {\scriptsize[36.0, 38.0]} & \textbf{89.1} {\scriptsize$\pm$2.9} & 46.0 {\scriptsize$\pm$3.9} & \textbf{3.03} {\scriptsize$\pm$0.39}M \\
    \addlinespace[2pt]
    \rowcolor[gray]{0.93} \multicolumn{5}{c}{\emph{Implementer:} \textbf{GPT-5.6 Luna}} \\
    \quad Astra & 44.3 {\scriptsize[43.4, 45.2]} & 85.3 {\scriptsize$\pm$1.8} & 57.3 {\scriptsize$\pm$1.7} & 3.00 {\scriptsize$\pm$0.22}M \\
    \quad GPT-6 Sol & 36.8 {\scriptsize[35.9, 37.7]} & 83.4 {\scriptsize$\pm$2.1} & 49.3 {\scriptsize$\pm$1.5} & 2.24 {\scriptsize$\pm$0.19}M \\
    \quad Fable & 46.2 {\scriptsize[45.3, 47.2]} & 85.6 {\scriptsize$\pm$1.6} & 60.8 {\scriptsize$\pm$1.9} & 3.64 {\scriptsize$\pm$0.27}M \\
    \quad Opus 5.5 & \textbf{47.3} {\scriptsize[46.3, 48.3]} & 85.9 {\scriptsize$\pm$1.5} & \textbf{62.1} {\scriptsize$\pm$2.0} & 3.35 {\scriptsize$\pm$0.24}M \\
    \quad GLM 5.3 & 40.5 {\scriptsize[39.5, 41.5]} & 83.5 {\scriptsize$\pm$1.8} & 55.3 {\scriptsize$\pm$1.7} & 3.93 {\scriptsize$\pm$0.26}M \\
    \quad Grok & 39.0 {\scriptsize[37.7, 40.4]} & 83.1 {\scriptsize$\pm$1.9} & 53.1 {\scriptsize$\pm$1.7} & 3.22 {\scriptsize$\pm$0.21}M \\
    \quad DeepSeek V4 Pro & 31.2 {\scriptsize[30.1, 32.4]} & 83.1 {\scriptsize$\pm$1.8} & 42.6 {\scriptsize$\pm$1.2} & 2.79 {\scriptsize$\pm$0.18}M \\
    \quad Kimi K3 & 42.6 {\scriptsize[41.2, 43.9]} & 84.7 {\scriptsize$\pm$1.6} & 57.1 {\scriptsize$\pm$1.8} & 2.99 {\scriptsize$\pm$0.20}M \\
    \cmidrule(lr){1-5}
    \quad Production library & 45.5 {\scriptsize[44.6, 46.3]} & 84.5 {\scriptsize$\pm$2.7} & 60.4 {\scriptsize$\pm$2.8} & 4.26 {\scriptsize$\pm$0.73}M \\
    \quad No library & 33.3 {\scriptsize[32.6, 33.9]} & \textbf{86.1} {\scriptsize$\pm$3.0} & 44.3 {\scriptsize$\pm$2.6} & \textbf{1.46} {\scriptsize$\pm$0.31}M \\
    \addlinespace[2pt]
    \rowcolor[gray]{0.93} \multicolumn{5}{c}{\emph{Implementer:} \textbf{GLM 5.3 Flash}} \\
    \quad Astra & 44.1 {\scriptsize[42.7, 45.4]} & 84.5 {\scriptsize$\pm$1.8} & 60.5 {\scriptsize$\pm$1.7} & 5.74 {\scriptsize$\pm$0.42}M \\
    \quad GPT-6 Sol & 37.3 {\scriptsize[36.0, 38.6]} & 85.3 {\scriptsize$\pm$1.8} & 54.3 {\scriptsize$\pm$1.3} & 5.46 {\scriptsize$\pm$0.40}M \\
    \quad Fable & 45.9 {\scriptsize[44.4, 47.4]} & 84.7 {\scriptsize$\pm$1.7} & 64.2 {\scriptsize$\pm$1.9} & 8.13 {\scriptsize$\pm$0.74}M \\
    \quad Opus 5.5 & \textbf{47.9} {\scriptsize[46.4, 49.3]} & 85.5 {\scriptsize$\pm$1.5} & \textbf{67.2} {\scriptsize$\pm$2.0} & 7.13 {\scriptsize$\pm$0.60}M \\
    \quad GLM 5.3 & 40.0 {\scriptsize[38.4, 41.5]} & 82.6 {\scriptsize$\pm$1.8} & 58.7 {\scriptsize$\pm$1.9} & 9.06 {\scriptsize$\pm$0.69}M \\
    \quad Grok & 37.9 {\scriptsize[36.2, 39.6]} & 84.0 {\scriptsize$\pm$1.6} & 55.4 {\scriptsize$\pm$1.8} & 7.88 {\scriptsize$\pm$0.58}M \\
    \quad DeepSeek V4 Pro & 29.1 {\scriptsize[27.8, 30.4]} & 84.0 {\scriptsize$\pm$1.7} & 42.3 {\scriptsize$\pm$1.2} & 6.33 {\scriptsize$\pm$0.46}M \\
    \quad Kimi K3 & 43.0 {\scriptsize[41.1, 44.9]} & 85.4 {\scriptsize$\pm$1.7} & 61.0 {\scriptsize$\pm$1.9} & 6.69 {\scriptsize$\pm$0.51}M \\
    \cmidrule(lr){1-5}
    \quad Production library & 45.1 {\scriptsize[44.0, 46.2]} & \textbf{85.6} {\scriptsize$\pm$2.6} & 62.1 {\scriptsize$\pm$2.8} & 9.83 {\scriptsize$\pm$1.90}M \\
    \quad No library & 32.9 {\scriptsize[31.7, 34.0]} & 83.7 {\scriptsize$\pm$2.7} & 48.7 {\scriptsize$\pm$3.1} & \textbf{3.28} {\scriptsize$\pm$0.66}M \\
    \bottomrule
    \end{tabular}

\end{table}

\begin{figure*}[h]
  \centering
  \includegraphics[width=1\linewidth]{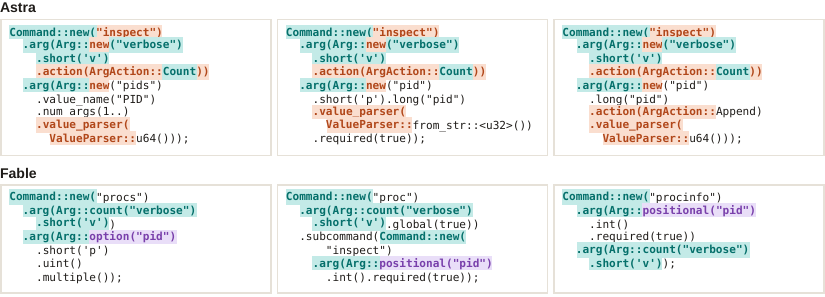}
  \caption{\textbf{Agents converge on the same designs.} README quick-starts from Astra (top) and Fable (bottom) across six \texttt{clirs} libraries each. \hlshared{Teal}: all six; \hlastra{orange}: Astra only; \hlfable{violet}: Fable only.}
  \label{fig:impl-problems}
\end{figure*}

\paragraph{Agents reproduce production-library abstractions.} Designers converge on the same design in eleven of fifteen tasks. We assessed this by inspecting the libraries, with agent verification. In \clirs{}, Astra and Fable copy \texttt{clap}'s builder design, not its shorter derive macro (\autoref{fig:impl-problems}).

\paragraph{Performance by implementer.} \autoref{tab:arm-ladder} shows that all three implementers agree on the top three designers (Opus 5.5, Fable 5.1, GPT-6 Astra) and the last (DeepSeek V4 Pro), and none favors its own model family. Absolute scores differ across implementers, and averaging weights each equally.

\subsection{Failure Taxonomy}\label{sec:failure-taxonomy}
\begin{figure*}[h]
  \centering
  \includegraphics[width=1\linewidth]{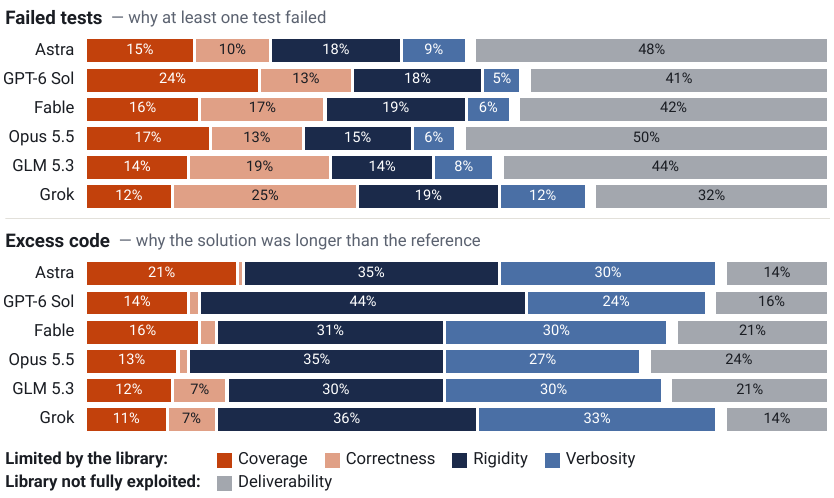}
  \caption{Primary failure category per designer. \textbf{Failed Tests} classifies why at least one test failed. \textbf{Excess Code} classifies why the solution was longer than the reference. Definitions are in \autoref{tab:taxonomy-definitions}.}
  \label{fig:tax-counts}
\end{figure*}

Agent-written libraries resemble production libraries, yet implementers still write more code than the reference. We audit \taxexcesscases{} partially passing solutions that exceed their references in code size, sampled from six designer configurations. We classify each failure by the smallest library change sufficient to prevent it. \autoref{appendix:taxonomy} gives the procedure and scope. The categories are \textit{Coverage} (nothing close to the needed capability exists), \textit{Correctness} (the capability exists but has a bug), \textit{Rigidity} (it nearly fits but cannot be adapted to the task), \textit{Verbosity} (it fits but requires excess code), and \textit{Deliverability} (a simpler path exists, but the implementer did not find it).

In this sample, \taxexcessatceilingpct{} of primary excess-code classifications fall under library limitations (\autoref{fig:tax-counts}). Rigidity and Verbosity account for \taxexcessergonomicspct{}, compared with \taxexcesscoveragepct{} for Coverage. A common pattern we observe is that agent-written libraries implement only the exact core functionality the specification names. In \texttt{clirs}, 24 of the 33 written libraries do not easily implement long-option prefix inference, a capability expected under our full-featured-library brief. On problems that need it, libraries lacking it score 16.7 points lower than those that provide it. In an author analysis of agent-collected evidence, separate from the \taxexcesscases{}-cell audit, 41\% of 13,562 hand-labeled lines that implementers rebuilt in 180 solutions redo functionality the library shipped but hid or broke.

\subsection{How Do Agents Use Libraries?}\label{sec:lib-usage}
\begin{figure}[t]
  \centering
  \begin{minipage}[t]{0.49\textwidth}
    \vspace{0pt}
    \centering
    \captionsetup{position=top}
    \captionof{figure}{\textbf{LibraryUseBench score vs.\ cost per problem.} All models run in mini-SWE-agent. Bars are 95\% intervals.}
    \label{fig:library-use-bench}
    \includegraphics[width=\linewidth]{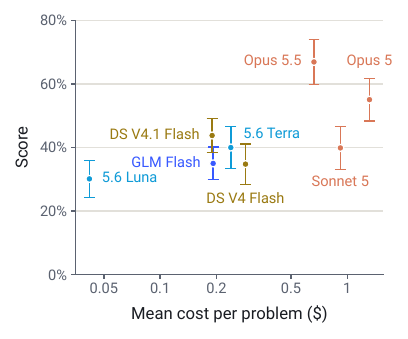}
  \end{minipage}\hfill
  \begin{minipage}[t]{0.49\textwidth}
    \vspace{0pt}
    \centering
    \captionof{table}{\textbf{GPT-5.6 Luna (Codex, High) results with different reasoning efforts and prompts}. Effort rows use the default prompt, and prompt rows use high effort.}
    \label{tab:library-use-levers}
    {\footnotesize\setlength{\tabcolsep}{2.5pt}    \begin{tabular}{llrrrr}
    \toprule
    Lever & Setting & Score & \% Pass & Simplicity & \$/prob. \\
    \midrule
    Effort & Low & 27.8 & 51.4 & \textbf{83.2} & \textbf{\$0.023} \\
     & Med. & 38.9 & 69.6 & 72.5 & \$0.055 \\
    \midrule
    Prompt & Min. & 37.0 & 85.9 & 48.7 & \$0.082 \\
     & Low & 36.5 & \textbf{87.3} & 46.4 & \$0.105 \\
     & Med. & 44.3 & 85.7 & 57.3 & \$0.111 \\
    \midrule
    Default & High & \textbf{45.5} & 84.5 & 60.4 & \$0.149 \\
    \bottomrule
    \end{tabular}
}
  \end{minipage}
\end{figure}

LibraryUseBench measures how well agents use production libraries. Each of \lubmodels{} models gets the production library and the minimal prompt (\autoref{lst:prompt-minimal}), which only asks it to use the library and write little code. The GPT-5.6 Luna prompt and reasoning-effort comparisons in Table~\ref{tab:library-use-levers} instead use Codex. \autoref{fig:library-use-bench} shows that natural library use scales with model capability.

Looking deeper, we vary prompt prescription and reasoning effort for GPT-5.6 Luna (Table~\ref{tab:library-use-levers}). The default prompt is given in \autoref{lst:main-prompt} and its variants in \autoref{appendix:lib-usage}. Raising effort from low to high increases the pass rate from 51.4\% to 84.5\%, with Luna reading 15 library files instead of 2. The default prompt instead raises simplicity from 48.7 to 60.4 over the minimal one. Under the minimal prompt, 24\% of trials never open the library and 36\% of the reference's library symbols go unseen, versus 5\% under the default. Luna searches by grepping API names it already expects, so capabilities it does not recall go unused (\autoref{appendix:lib-search}). Even at the default setting, its simplicity remains well below the reference's.

\subsection{Do Agentic Design Patterns Work?}\label{sec:agent-preferences}
\begin{figure*}[h]
  \centering
  \includegraphics[width=1\linewidth]{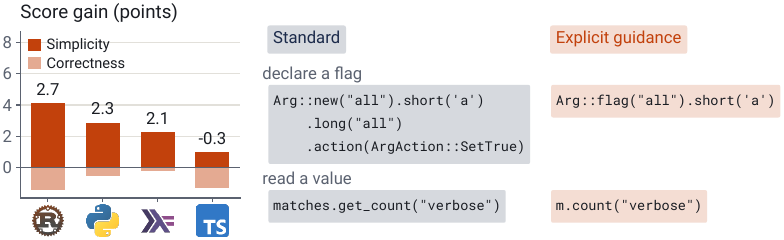}
  \caption{\textbf{Explicit guidance results}. Left: score gain per language, split into simplicity and correctness contributions (\autoref{appendix:neuralese}). Right: \texttt{clirs} examples under each prompt.}
  \label{fig:neuralese}
\end{figure*}

We next test whether more prescriptive agent-first guidance changes the resulting libraries. GPT-6 Astra (Codex, high) designs each library with and without an explicit guidance prompt appended to the unchanged specification (\autoref{appendix:neuralese}). The prompt states that only coding agents, scored on how little code they write, will use the library; it provides design guidance intended for agent users and combines consumer-first API sketches, runnable usage examples, and testing with subagents. We evaluate the combined intervention, not each component separately.

Guided libraries share fewer exported names with production, \neuraleseincumbentrecall\% versus \standardincumbentrecall\% (\autoref{appendix:neuralese}). In \clirs{}, builder chains become single task-level calls (\autoref{fig:neuralese}, right). The score rises from \neuralesestandardscore{} to \neuralesescore{}, just below production (\existinglibrarymeanscore{}) and standard-prompt \sotamodel{} (\sotascore{}). The score improves for all three implementers and three of four languages (\autoref{fig:neuralese}, left), mostly via shorter solutions. Guidance nearly doubles design cost (\neuraleseauthorcost{} vs.\ \neuralesestandardauthorcost{}) with unchanged per-problem cost ($\neuralesecostchangepm$).

\section{Related Work}
\textbf{Agents that write reusable code.} Agents generate repositories~\citep{nl2repo,projecteval,silin_chen_repo0_2026,zhaoxi_zhang_repozero_2026,jiale_zhao_denovoswe_2026,commit0,ruida_hu_evaluating_2026,icae,john_yang_programbench_2026}, refactor code into libraries~\citep{lilo,regal,ziga_kovacic_refactoring_2025,refactorbench,shapelib,codetaste}, work over long horizons~\citep{scb,clb,huang2026deepswe,seniorswe,cognition2026frontiercode,frontierswe}, and write tools for themselves or weaker models~\citep{latm,creator,craft,trove}. Most are graded by tests or task accuracy, which cannot see design. Agents can pass nearly every test while leaving the requested library unused~\citep{buildingtothetest}. The closest, ReGAL~\citep{regal} and LATM~\citep{latm}, pass one model's code to another, but build small functions from solved tasks rather than design a library from an open specification. \ldb{} grades design by downstream correctness and simplicity.

\textbf{Evaluating libraries and their use.} Software engineering judges an API by observing developers on fixed tasks, sometimes under competing designs~\citep{ellis2007factory,EmpiricalStudyAPI2013,rauf2019apiusability}, which finds that both API structure and documentation hinder developers~\citep{robillard2009apis,myers2016apiusability}, or by scoring the API surface~\citep{scheller2015apiconcepts}, the code that reuse saves~\citep{frakes1996reuse}, or complexity metrics~\citep{mcabecyc,Halstead1977ElementsOS,cogcomplex} that need not reflect what LLMs find hard~\citep{chen_xie_rethinking_2026,shaswat_patel_is_2026}. LLM benchmarks fix a library and score whether models call it correctly~\citep{ds1000,bigcodebench,zan2022private,gorilla,chong_wang_llms_2024,nihal_jain_cloudapibench_2024,jingyi_chen_when_2025}. \ldb{} keeps the user-study design with agents as users, and replaces completion time with correctness and code size against an expert reference. LibraryUseBench applies this measure to production libraries.

\textbf{Designing for agents.} Recent work studies how LLMs write and use code~\citep{llmimpactsoftware,speequalitycursor,twist-etal-2026-study,lukas_twist_study_2024,kan_watanabe_what_2026} and argues code should be designed with agents as consumers~\citep{shaolin_wang_human_2026,borg_code_2026,shaswat_patel_is_2026}, and agent-specific interfaces do help agents~\citep{minisweagent}. Evaluations fix the consumer and vary its framework or documentation~\citep{jintao_huang_adk_2026,sandya_wijaya_readmellm_2025}, score agent-built tools one at a time~\citep{kaliyev_beyond_2026}, or grade the agents that coding agents build~\citep{quan_shi_tautau-bench_2026}. None varies the designer. To our knowledge, \ldb{} is the first to measure how well an agent designs a library for other agents.

\section{Limitations}\label{sec:limitations}

\ldb{} evaluates the downstream value of a library under specified tasks, consumer configurations, and execution budgets. Scores characterize utility in that setting, not a consumer-independent library ordering. It measures tested correctness and the static size and complexity of consumer programs, not full library correctness, security, runtime efficiency, or maintainability. Reference programs normalize scores without prescribing the generated API, but are not uniquely optimal. The benchmark emphasizes workloads with opportunities for reuse, and its confidence intervals cover reruns of the fixed task set, not generalization to all library domains.

\section{Conclusion}

We introduced \ldb{}, a benchmark that evaluates an agent-designed library only through how downstream agents use it, scoring the correctness and simplicity of their programs against expert references built on real production libraries. Across fifteen tasks in four languages, agents design libraries that help other agents, with \sotamodel{} scoring \sotascoreimprovepct{}\% above the production library, but they do so by reproducing the production library's design in eleven of fifteen tasks. Downstream agents then underuse these libraries. Most excess code comes from rigid or hard-to-use interfaces rather than missing capabilities. Telling designers that only agents will use their library and having them test it with subagents yields libraries that copy fewer production designs and shrink downstream programs, yet the result still falls just short of the production library. Designing libraries for agents is therefore not the same as designing them for humans, and it remains an open problem. We hope \ldb{} serves as a testbed for studying which interfaces, abstractions, and documentation help agents build on one another's code.

\subsubsection*{Acknowledgments} %

We would like to thank John Yang, Parth Asawa, Xavier Garcia, Ryan Carelli, Arun Kumar, Florian Brand, and Nick Roberts for their helpful feedback and discussions. This work was supported by the Snorkel AI Open Benchmark grant, DARPA, the NSF, and by the Prime Intellect residency.

\bibliography{iclr2027_conference}
\bibliographystyle{iclr2027_conference}

\appendix
\raggedbottom
\section*{Appendix}
\section{Static Measurement Details}\label{appendix:staticmeasures}

\paragraph{Pipeline.}
We first format each submission with a pinned formatter set to 80 columns, so all code has the same layout and line counts are fair. Code that fails to format scores zero. We then parse each source file with tree-sitter (0.25.2) and count every metric in one pass over the syntax tree. Build output, dependencies, and symlinks are skipped. If the parser cannot read part of a file, we still measure the rest. The formatter decides whether code is valid.

\paragraph{Metrics.}
\begin{itemize}
  \item \textbf{SLOC}: nonblank lines, not counting comments or Python
    docstrings.
  \item \textbf{Cyclomatic complexity}: $+1$ for each function, each branch,
    and each \texttt{\&\&} or \texttt{||}.
  \item \textbf{Cognitive complexity}: each branch adds $1$ plus its nesting
    depth, and code inside it is one level deeper. Each \texttt{\&\&} or
    \texttt{||} adds $1$. Functions do not add depth.
  \item \textbf{Halstead volume}: $N \log_2 \eta$, where $N$ is the number of
    operators and operands and $\eta$ is the number of distinct ones across the
    whole workspace. Operands are names and literals.
\end{itemize}

\paragraph{Python} (Ruff 0.16.6).
\texttt{if}, \texttt{for}, \texttt{while}, \texttt{except}, ternaries, and the
\texttt{for}/\texttt{if} parts of comprehensions are branches. \texttt{elif}
adds $1$ to each complexity, with no nesting cost. \texttt{match} adds only to
cognitive complexity, and each \texttt{case} except \texttt{\_} adds $1$ to
cyclomatic. \texttt{assert} adds $1$ to cyclomatic. A \texttt{lambda} does not
count as a function.

\paragraph{JavaScript and TypeScript} (Prettier 3.9.6).
Both use the same rules. \texttt{if}, loops, \texttt{catch}, each
\texttt{switch} case, and ternaries are branches. An \texttt{else if} counts as
an \texttt{if} nested inside the one before it, so long chains cost more. Every
function counts, including arrow functions and methods. \texttt{??} adds $1$ to
cyclomatic only. A TypeScript submission also counts JavaScript files it uses.

\paragraph{Rust} (rustfmt 1.9.0).
\texttt{if}, \texttt{for}, \texttt{while}, \texttt{loop}, and \texttt{match}
are branches. A \texttt{match} counts once; its arms add nothing. A labeled
\texttt{break} or \texttt{continue} adds $1$ to cognitive. A function that calls
itself adds $1$ to cognitive, once. Code inside macros and the \texttt{?}
operator counts as zero.

\paragraph{Haskell} (Fourmolu 0.20.1.0).
Each equation of a function adds $1$ to cyclomatic. \texttt{if} is a branch.
A \texttt{case} adds to cognitive like a branch and adds $1$ to cyclomatic for
each alternative after the first. Guards work like an if/else-if chain. Each
guard adds $1$ to cyclomatic. The first guard adds $1$ plus nesting depth to
cognitive, and each later guard, including \texttt{otherwise}, adds $1$.
\texttt{<|>} counts like \texttt{||}.

\section{Setup Details}\label{appendix:setup}

\paragraph{Agents.} All agents run in mini-SWE-agent~\citep{mswea} unless otherwise specified, at ``High'' reasoning effort. The designers are GPT-5.6 Sol (Codex)~\citep{gpt56}, GPT-6 Sol~\citep{gpt6sol}, GPT-6 Astra~\citep{gpt6astra}, Opus 5.5~\citep{opus55}, Fable 5.1~\citep{fable51}, GLM 5.3~\citep{glm53}, Grok 4.6~\citep{grok46}, DeepSeek V4 Pro~\citep{deepseekv4}, and Kimi K3~\citep{kimik3}. Fable 5.1 also runs in Claude Code, abbreviated CC, and GPT-6 Astra in Codex. The implementers are GPT-5.6 Luna (Codex)~\citep{gpt56}, GLM 5.3 Flash~\citep{glm53}, and DeepSeek V4.1 Flash~\citep{deepseekv4}, each making one attempt per problem with the same prompt. LibraryUseBench (\autoref{sec:lib-usage}) evaluates these three implementer models plus Opus 5.5~\citep{opus55}, Opus 5~\citep{opus5}, Sonnet 5~\citep{sonnet5}, GPT-5.6 Terra~\citep{gpt56}, and DeepSeek V4 Flash~\citep{deepseekv4} as implementers on the production library, all evaluated in mini-SWE-agent. The GPT-5.6 Luna prompt and reasoning-effort comparisons (Table~\ref{tab:library-use-levers}) instead use Codex. OpenAI, Anthropic, and xAI models use first-party APIs, GLM uses OpenRouter, and DeepSeek and Kimi use Prime Inference. Costs use official list prices as of September 2026.

\paragraph{Environment.} Harbor~\citep{harbor} runs one Docker image per task, shared by both phases and every library condition (\autoref{tab:shared-environment}). Preinstalled packages are not target-domain libraries, and health checks block substituting a human-written library. The library is mounted at \texttt{/library} and the agent works in \texttt{/workspace}. The agent network is limited to model-provider APIs. Library specifications get packaging instructions appended (\autoref{sec:packaging}). Limits are in \autoref{tab:limits}, and pinned formatters are in \autoref{tab:formatters}.

\begin{table}[ht]
     \centering
     \small
     \caption{Per-task Docker images, shared by both phases and all library conditions. Counts are tasks.}
     \label{tab:shared-environment}
     \begin{tabular}{p{0.18\linewidth}p{0.35\linewidth}p{0.35\linewidth}}
       \hline
       Scope & Toolchain & Dependencies \\
       \hline
       All (15) & \texttt{bash}, \texttt{git}, \texttt{curl}, \texttt{jq}, \texttt{rg}, Node 22.14, Python 3.12, \texttt{uv} 0.9.5 & -- \\
       Python (6) & \texttt{ruff} & Pinned \texttt{requirements.txt} in \texttt{/workspace/.venv}; \texttt{uv} cache warmed \\
       TypeScript (1) & \texttt{tsc}, \texttt{tsx}, \texttt{prettier}, Chromium & Pinned \texttt{package.json} \\
       Rust (5) & \texttt{cargo}, \texttt{rustfmt}, \texttt{clippy}; Rust 1.85--1.91 & Prefetched \texttt{Cargo.lock}; \texttt{CARGO\_NET\_OFFLINE} \\
       Haskell (3) & GHC 9.8.4, \texttt{cabal}, \texttt{fourmolu} & Prefetched frozen Cabal closure \\
       \hline
     \end{tabular}
\end{table}

\begin{table}[ht]
     \centering
     \small
     \caption{Harness-enforced limits per phase. The agent budget is wall-clock inside the
     container. Design runs against wall-clock alone; a problem ends at
     whichever of its two budgets binds first.}
     \label{tab:limits}
     \begin{tabular}{p{0.26\linewidth}p{0.30\linewidth}p{0.30\linewidth}}
       \hline
        & \phaseone{} (design) & \phasetwo{} (one problem) \\
       \hline
       Agent wall-clock & 4 hours & 60 minutes \\
       Verifier wall-clock & 15 minutes & 5--60 minutes, set per problem \\
       Sandbox & 4 CPU, 8\,GB & 2 CPU, 4\,GB \\
       Spend cap & None & \$2.50 \\
       Attempts & 1 ($\numruns=3$ independent runs per task) & 1 \\
       Agent network & Model-provider API allowlist only & Model-provider API allowlist only \\
       Verifier network & Disabled & Disabled \\
       \hline
     \end{tabular}
\end{table}

\begin{table}[ht]
     \centering
     \small
     \caption{Pinned normalizers and grammars used for static measurement. The same
     versions are applied to the optimized references and to every generated program.
     Measurement uses its own pinned Rust 1.98.1 toolchain for \texttt{rustfmt}, separate
     from the per-task build toolchains in \autoref{tab:shared-environment}; TypeScript problems
     additionally load the JavaScript grammar for embedded sources.}
     \label{tab:formatters}
     \begin{tabular}{p{0.13\linewidth}p{0.30\linewidth}p{0.21\linewidth}p{0.18\linewidth}}
       \hline
       Language & Formatter (pinned) & Config & \texttt{tree-sitter} grammar \\
       \hline
       Python & \texttt{ruff format} 0.16.6 & \texttt{ruff.toml} & \texttt{python} \\
       TypeScript & \texttt{prettier} 3.9.6 & \texttt{prettier.json} & \texttt{typescript} \\
       Rust & \texttt{rustfmt} 1.9.0-stable & \texttt{rustfmt.toml} & \texttt{rust} \\
       Haskell & \texttt{fourmolu} 0.20.1.0 & \texttt{fourmolu.yaml} & \texttt{haskell} \\
       \hline
     \end{tabular}
\end{table}
\section{Library Packaging Instructions}\label{sec:packaging}

Every \phaseone{} specification ends with a general-instructions block appended
to the capability brief in \autoref{fig:hero}. The block asks for all
functionality users would reasonably expect of such a library, states that it
will be used primarily by coding agents and the senior engineers who review
their work, and fixes a language-specific package layout so the \phasetwo{}
harness can install the library without guessing. Each layout names the package
manager and package name and requires a lockfile limited to the offline
dependencies the block lists. Where a single build command exists, the block
states it and the library must pass it offline.

\begin{itemize}
  \item \textbf{Python.} \texttt{pyproject.toml} with \texttt{[project].name},
    managed by \texttt{uv}, with a \texttt{uv.lock}.
  \item \textbf{TypeScript.} \texttt{package.json} with \texttt{name} and a
    \texttt{package-lock.json}. The package ships the runtime files its public
    entry points need and TypeScript declarations for its public interface.
  \item \textbf{Rust.} \texttt{Cargo.toml} with \texttt{[package].name} and a
    \texttt{Cargo.lock}. It builds with
    \texttt{cargo build --manifest-path /workspace/Cargo.toml --offline}.
  \item \textbf{Haskell.} A top-level \texttt{.cabal} file naming the library and
    exposing its modules, plus \texttt{cabal.project} and
    \texttt{cabal.project.freeze}. It builds with \texttt{cabal build all --offline}
    and as a source dependency of another Cabal project.
\end{itemize}

The block does not constrain the public interface, the module decomposition, or
the abstractions. That design freedom is what \ldb{} measures.

\section{Main Prompts Used}\label{sec:prompts}

\noindent\begin{minipage}{\linewidth}
\begin{lstlisting}[style=lstprompt,caption={Implementer prompt.},label={lst:main-prompt}]
<task instruction>

## Library Rules

Your solution is a thin adapter around `<library>`. It is judged on how
little code sits on top of the library, so every operation the library can
carry, the library carries.

- `<library>` is installed. Its source, examples, tutorials, and docs are in `/library` (read-only).
- Reach for the primitive that does the whole operation (the parser, validator, pipeline, runner), not its pieces. Importing constants, error types, or small helpers while hand-rolling the operation is not using the library.
- If the library's default behavior differs from the task, configure or extend the library. Reimplementing is the last resort, and only after a search confirms the library lacks it.
- Handle exactly the validation the task describes.
- Other available dependencies: <dependencies>. Use them for work outside the library's domain.
- You have one hour.

## Workflow

1. **Map the task onto the library.** Read `/library` in this order: README and docs, then examples, then grep the source for each concept the task names. Done when every requirement in the task is paired with the library entry point that carries it, or with "not provided" after a search.
2. **Build the program in `/workspace`** by calling those entry points.
3. **Audit.** For each function, loop, branch, and check you wrote, name the library call that replaces it and use that instead, or note why the library lacks it. Done when every remaining hand-written line has a reason.
4. **Run the task's sample inputs.** Done when each produces the described output. Sample runs are enough; skip test suites.
5. Submit.
\end{lstlisting}
\end{minipage}

\noindent\begin{minipage}{\linewidth}
\begin{lstlisting}[style=lstprompt,caption={No-library implementer prompt.},label={lst:prompt-nolib}]
<task instruction>

## Rules

Your solution is judged on how little code it takes, so keep it as small and
direct as the task allows.

- Available dependencies: <dependencies>. Use them for work outside the task's core domain (CLI parsing, serialization, etc.). Nothing else is installed.
- Handle exactly the validation the task describes.
- You have one hour.

## Workflow

1. **Build the program in `/workspace`.**
2. **Run the task's sample inputs.** Done when each produces the described output. Sample runs are enough; skip test suites.
3. Submit.
\end{lstlisting}
\end{minipage}

\section{Benchmark Problems}\label{sec:problems}

\autoref{tab:problems} lists the fifteen library-design problems, the
production library each one's reference solutions are written against, and the
number of \phasetwo{} problems built for it. The library name is the name the
designer agent is told to package under, and is the name used throughout this
paper.

\begin{table}[ht]
     \centering
     \small
     \caption{The fifteen \ldb{} library-design problems.}
     \label{tab:problems}
     \begin{tabular}{llll r}
       \hline
       Library & Language & Production library & Domain & Problems \\
       \hline
       \texttt{canon}      & Python     & \texttt{pydantic}        & Schema validation & 14 \\
       \texttt{pyda}       & Python     & \texttt{pandas}          & Dataframes & 21 \\
       \texttt{roadkill}   & Python     & \texttt{uxsim}           & Traffic simulation & 16 \\
       \texttt{sapi}       & Python     & \texttt{fastapi}         & HTTP API framework & 11 \\
       \texttt{simu}       & Python     & \texttt{simpy}           & Discrete-event simulation & 13 \\
       \texttt{uglypie}    & Python     & \texttt{beautifulsoup4}  & HTML parsing & 11 \\
       \texttt{heretical}  & TypeScript & \texttt{dompurify}       & HTML sanitization & 22 \\
       \texttt{arazu}      & Rust       & \texttt{rapier}          & Rigid-body physics & 22 \\
       \clirs{}            & Rust       & \texttt{clap}            & CLI construction & 13 \\
       \texttt{npr}        & Rust       & \texttt{ndarray}         & N-dimensional arrays & 20 \\
       \texttt{rubberband} & Rust       & \texttt{tantivy}         & Full-text search & 20 \\
       \texttt{weft}       & Rust       & \texttt{chumsky}         & Parser combinators & 14 \\
       \texttt{plaid}      & Haskell    & \texttt{megaparsec}      & Parser combinators & 15 \\
       \texttt{umami}      & Haskell    & \texttt{unagi-chan}      & Concurrent channels & 20 \\
       \texttt{wapi}       & Haskell    & \texttt{servant}         & Typed web APIs & 10 \\
       \hline
       \multicolumn{4}{l}{\textbf{Total}} & \textbf{242} \\
       \hline
     \end{tabular}
\end{table}

\section{Taxonomy}\label{appendix:taxonomy}
\paragraph{Audit protocol.} For each generated library from the \taxauthors{} mini-SWE-agent designers, we sample \taxcellsperlibrary{} downstream cells (one implementer's solution to one problem) that both failed at least one test and wrote more code than the reference. Every sampled cell also passed some tests, and cells that passed every test are not audited. One GPT-5.6 Luna agent with high reasoning audits each cell from the library, the implementer's trajectory and solution, the test results, and the reference solution. It describes the best as-shipped path without executing it and returns up to three causes per symptom, largest first. We report the first as the cell's primary classification, giving \taxexcesscases{} excess-code and \taxfailedcases{} failed-test classifications from the same cells.

\enlargethispage{2\baselineskip}
\paragraph{Scope.} Each cell is audited once by one model, from the same family as one implementer, without human labels or an agreement check. The auditor reasons about the best as-shipped path rather than running it. It attributes every failure to the library change that would have prevented it, including an implementer's own task-logic errors. The shares describe partially passing cells, not every downstream cell.

\begin{table*}[t]
  \centering
  \small
  \caption{Failure-taxonomy categories and leaves; \autoref{appendix:taxonomy} gives the classification procedure.}
  \label{tab:taxonomy-definitions}
  \begin{tabular}{@{}>{\raggedright\arraybackslash}p{0.15\linewidth}>{\raggedright\arraybackslash}p{0.22\linewidth}p{0.55\linewidth}@{}}
    \toprule
    Category & Leaf & Definition \\
    \midrule
    \multicolumn{3}{@{}l}{\emph{Limited by the library: no path through the library as shipped does better.}} \\
    \addlinespace
    Coverage & Absent operation & No operation or documented composition performs this reusable domain computation, and none does nearly this. The fix is a new operation. \\
    \addlinespace
    Correctness & Contract violation & The library returns wrong output on a legitimate input. \\
     & Misleading diagnostic & An error pointed away from the actual cause, and the implementer followed it. \\
     & Performance defect & The library path is too slow for the problem's limits. \\
    \addlinespace
    Rigidity & Fixed policy & An operation hard-codes how it works, such as ordering, rounding, error handling, or output format, with no parameter that selects what the problem needs. \\
     & Closed representation & The library's data model cannot represent the problem's value, field, or variant, so the implementer builds parallel types. \\
     & Monolithic operation & An operation bundles several steps; the implementer needs one of them or a small variant, and the inner steps are not exposed. \\
     & Excluded scope & A primitive excludes, by design, what the problem feeds it, such as a mode, encoding, or input layout. \\
    \addlinespace
    Verbosity & Verbose interface & Using the library takes at least as much code as writing the logic by hand. \\
     & Unpackaged composition & Reusable multi-step wiring around an operation that no helper packages. \\
     & Shape mismatch & Mechanical conversion between the library's inputs or outputs and the shapes the problem needs. \\
     & Repeated declaration & The same setting must be declared at several sites, with no shared place for it. \\
    \midrule
    \multicolumn{3}{@{}l}{\emph{Library not fully exploited: a path through the library as shipped removes the code or fixes the failure.}} \\
    \addlinespace
    Deliverability & Not surfaced & The implementer's reads never returned the capability, because of where it lives, what it is called, or what is exported. \\
     & Not recognizable & The implementer saw it, but its description does not express the need in the problem's terms. \\
     & Not salient & The implementer saw an adequate description, but it did not reach the decision: deep in a long output, cut off, or missed by a later search. \\
     & Incomplete contract & The implementer found the capability but not a fact needed to use it correctly, such as a precondition, default, or required setting. \\
     & Undistinguished alternatives & Several adjacent options were available, and nothing said which one fits. \\
    \bottomrule
  \end{tabular}
\end{table*}

\begin{table}[h]
     \centering
     \caption{Failure-taxonomy leaf counts per designer, over the standardized
    subset of \textbf{Failed tests} and \textbf{Excess code} cells classified in
    \autoref{sec:failure-taxonomy}. Each designer contributes 135 cells per symptom
    stratum. Leaf definitions are in \autoref{tab:taxonomy-definitions}.}
    \label{tab:taxonomy-counts}
     \resizebox{\textwidth}{!}{%
       \begin{tabular}{llrrrrrrrrrrrrrr}
\toprule
 & & \multicolumn{7}{c}{Failed tests} & \multicolumn{7}{c}{Excess code} \\
\cmidrule(lr){3-9} \cmidrule(lr){10-16}
Category & Leaf & Astra & GPT-6 Sol & Fable & Opus 5.5 & GLM 5.3 & Grok & All & Astra & GPT-6 Sol & Fable & Opus 5.5 & GLM 5.3 & Grok & All \\
\midrule
\multicolumn{16}{@{}l}{\emph{Limited by the library}} \\
Coverage & Absent operation & 20 & 32 & 21 & 23 & 19 & 16 & 131 & 28 & 19 & 21 & 17 & 16 & 15 & 116 \\
Correctness & Contract violation & 9 & 17 & 22 & 17 & 25 & 33 & 123 & 1 & 2 & 3 & 2 & 9 & 9 & 26 \\
 & Misleading diagnostic & 0 & 0 & 0 & 0 & 0 & 0 & 0 & 0 & 0 & 0 & 0 & 0 & 0 & 0 \\
 & Performance defect & 5 & 0 & 1 & 0 & 1 & 1 & 8 & 0 & 0 & 0 & 0 & 1 & 0 & 1 \\
Rigidity & Fixed policy & 17 & 18 & 19 & 17 & 14 & 19 & 104 & 38 & 41 & 31 & 32 & 32 & 41 & 215 \\
 & Closed representation & 3 & 2 & 3 & 0 & 0 & 1 & 9 & 0 & 1 & 2 & 1 & 0 & 1 & 5 \\
 & Monolithic operation & 0 & 1 & 0 & 1 & 2 & 2 & 6 & 3 & 6 & 4 & 5 & 1 & 4 & 23 \\
 & Excluded scope & 4 & 3 & 4 & 2 & 3 & 4 & 20 & 6 & 12 & 5 & 9 & 7 & 2 & 41 \\
Verbosity & Verbose interface & 3 & 2 & 2 & 3 & 4 & 5 & 19 & 9 & 10 & 12 & 9 & 10 & 11 & 61 \\
 & Unpackaged composition & 6 & 2 & 6 & 2 & 7 & 8 & 31 & 27 & 22 & 26 & 26 & 29 & 32 & 162 \\
 & Shape mismatch & 3 & 3 & 0 & 3 & 0 & 3 & 12 & 4 & 1 & 3 & 0 & 1 & 1 & 10 \\
 & Repeated declaration & 0 & 0 & 0 & 0 & 0 & 0 & 0 & 0 & 0 & 0 & 1 & 0 & 0 & 1 \\
\midrule
\multicolumn{16}{@{}l}{\emph{Library not fully exploited}} \\
Deliverability & Not surfaced & 0 & 0 & 0 & 1 & 0 & 1 & 2 & 0 & 0 & 0 & 0 & 1 & 0 & 1 \\
 & Not recognizable & 11 & 12 & 13 & 9 & 14 & 8 & 67 & 9 & 12 & 7 & 12 & 14 & 1 & 55 \\
 & Not salient & 17 & 4 & 12 & 9 & 12 & 7 & 61 & 6 & 3 & 13 & 7 & 6 & 6 & 41 \\
 & Incomplete contract & 35 & 35 & 28 & 37 & 29 & 24 & 188 & 0 & 4 & 3 & 3 & 3 & 6 & 19 \\
 & Undistinguished alternatives & 2 & 4 & 4 & 11 & 5 & 3 & 29 & 4 & 2 & 5 & 11 & 5 & 6 & 33 \\
\midrule
Limited by the library & & 70 & 80 & 78 & 68 & 75 & 92 & 463 & 116 & 114 & 107 & 102 & 106 & 116 & 661 \\
Library not fully exploited & & 65 & 55 & 57 & 67 & 60 & 43 & 347 & 19 & 21 & 28 & 33 & 29 & 19 & 149 \\
Cases & & 135 & 135 & 135 & 135 & 135 & 135 & 810 & 135 & 135 & 135 & 135 & 135 & 135 & 810 \\
\bottomrule
\end{tabular}
     }
\end{table}

\pagebreak
\section{Library Usage Prompts}\label{appendix:lib-usage}
\lstdefinestyle{lstprompt}{basicstyle=\ttfamily\footnotesize,breaklines=true,breakatwhitespace=false,
  columns=fullflexible,keepspaces=true,frame=single,framerule=0.3pt,xleftmargin=0pt,
  breakindent=0pt,escapeinside={(*@}{@*)},literate={`}{\textasciigrave}1}

\noindent\begin{minipage}{\linewidth}
\begin{lstlisting}[style=lstprompt,caption={Minimal-prescription implementer prompt.},label={lst:prompt-minimal}]
<task instruction>

## General Instructions

- You **must** use `<library>` (Source is at `/library`, installed for you already).
- Write as little code as possible.
- These are all of the dependencies available to you: <dependencies>.
- You have one hour.
\end{lstlisting}
\end{minipage}

\noindent\begin{minipage}{\linewidth}
\begin{lstlisting}[style=lstprompt,caption={Low-prescription implementer prompt.},label={lst:prompt-low}]
<task instruction>

## General Instructions

- You **must** use `<library>` as much as possible in your solution to minimize its size.
    - `<library>` has already been installed for you. Its raw source, examples, tutorials, and docs are in `/library`.
- These are all of the dependencies available to you: <dependencies>.
- Your programs only need to handle what is described above.
- You have one hour.

**Suggested Workflow:**
1. Read the examples/tutorials/docs/etc in `/library` before writing code.
2. Build the program in `/workspace`.
3. Run it with examples to ensure it works. You do not need to write test suites.
4. Submit.
\end{lstlisting}
\end{minipage}

\noindent\begin{minipage}{\linewidth}
\begin{lstlisting}[style=lstprompt,caption={Medium-prescription implementer prompt.},label={lst:prompt-medium}]
<task instruction>

## General Instructions

- Your solution is a thin adapter around `<library>`. Every line you write by hand that `<library>` could have carried makes the solution worse.
    - `<library>` has already been installed for you. Its raw source, examples, tutorials, and docs are in `/library`. Treat it as read-only.
    - Use the highest-level interface of `<library>` that fits the task. Importing constants, error types, or small utilities does not count as using the library when it exposes a broader primitive for the same work.
    - Search `/library` before choosing an interface. Only hand-write operations you have confirmed are not provided by `<library>`.
- These are all of the dependencies available to you: <dependencies>. Use them for work outside the library's domain.
- Your programs only need to handle what is described above. Only handle the validation described!
- You have one hour.

**Suggested Workflow:**
1. Read the examples/tutorials/docs/etc in `/library` before writing code.
2. Build the program in `/workspace`.
3. Reread every function, loop, branch, and check you wrote and ask whether `<library>` already does it. If it does, delete your version and call the library. Ideally this finds nothing because you built on the library from the start.
4. Run it with sample inputs to ensure it works. You do not need to write test suites.
5. Submit.
\end{lstlisting}
\end{minipage}

\section{How Agents Search Libraries}\label{appendix:lib-search}
We classify every library-touching shell command in each LibraryUseBench and GPT-5.6 Luna trajectory as a search (a grep-style search over the library), documentation read, example read, source read, listing, or introspection call, and parse each solution's library imports. This covers 14 runs, 10{,}164 trials, and 412{,}802 shell commands. Command categories agree with manual labels on 59 of 60 spot-checked commands, and import extraction on 30 of 30. We count a reference-solution symbol as \emph{seen} when it appears in the output of a library read, which makes seen rates an upper bound.

\paragraph{Agents verify the API they remember.} Under the minimal prompt, 24\% of Luna trials never touch the library and another 24\% only list it or print its version. Every skip is on a well-known library (e.g., 85\% of \texttt{bs4} and 76\% of \texttt{pandas} trials), and 92\% of those solutions still import it from memory. When agents search, 64--87\% of grep patterns are identifier-shaped (e.g., \texttt{RevoluteJointBuilder}, \texttt{fn map\_with}), and only 16--20\% share a word with the task. The grep$\to$source$\to$grep loop appears in 38--80\% of trials in every run. Across LibraryUseBench models the loop is the same and only its volume differs. Opus 5.5 reads 1.4 library files per trial, while DeepSeek V4.1 Flash reads 4.9.

\paragraph{Prescription moves reading up front.} \autoref{tab:lib-search-prescription} shows that more prescriptive prompts add documentation reading before the first write and read more of the library, while the share of grep and source commands stays flat. Symbols the agent never saw become symbols it uses; seen-but-unused symbols stay at 17--22\% from the low prompt upward. The audit step of the default prompt is mostly stated rather than performed. After the first clean run, 56\% of trials mention an audit in their reasoning, but only 40\% read the library again, and fewer than 11\% read it within three commands of a failing run.

\begin{table}[h]
\centering\footnotesize
\caption{GPT-5.6 Luna library interaction by prescription level (high effort).}
\label{tab:lib-search-prescription}
\begin{tabular}{lrrrr}
\toprule
 & Minimal & Low & Medium & High (default) \\
\midrule
Never touch library (\%) & 23.7 & 0.0 & 0.0 & 0.1 \\
Docs/examples read before first write (\%) & 13 & 88 & 87 & 100 \\
Library grep before first write (\%) & 44 & 88 & 98 & 100 \\
Distinct library files read & 3.5 & 7.5 & 10.0 & 14.8 \\
Distinct grep terms & 12 & 20 & 33 & 46 \\
Grep share of library commands & .32 & .27 & .33 & .29 \\
Source share of library commands & .30 & .25 & .32 & .30 \\
\midrule
Reference symbols used (\%) & 55 & 64 & 74 & 78 \\
Reference symbols seen, not used (\%) & 9 & 22 & 18 & 17 \\
Reference symbols never seen (\%) & 36 & 14 & 8 & 5 \\
\midrule
Audit language after first clean run (\%) & 4 & 3 & 21 & 56 \\
Library read after first clean run (\%) & 25 & 25 & 35 & 40 \\
\bottomrule
\end{tabular}
\end{table}

\paragraph{Reasoning effort sets search depth.} \autoref{tab:lib-search-effort} shows that low effort stops at listings and documentation. It also concludes early that the library lacks a capability. Of the first ``library lacks X'' claims, 88--91\% come before the first write, and in a 150-trial hand-labeled sample only 22--38\% follow a grep for that capability. In one \texttt{tantivy} problem, a low-effort grep for \texttt{Stemmer} piped through \texttt{head -100} cut off the tokenizer registrations, and the agent concluded that stemming is not built in. It scored 0, while all six medium- and high-effort attempts used the default \texttt{en\_stem} tokenizer and scored 79--96.

\begin{table}[h]
\centering\footnotesize
\caption{GPT-5.6 Luna library interaction by reasoning effort (default prompt).}
\label{tab:lib-search-effort}
\begin{tabular}{lrrr}
\toprule
 & Low & Medium & High \\
\midrule
Distinct library files read & 2.0 & 5.7 & 14.8 \\
Library commands per trial & 3.1 & 5.3 & 12.2 \\
Any library grep (\%) & 45 & 60 & 91 \\
Any library source read (\%) & 23 & 50 & 81 \\
Only listings or docs (\%) & 42 & 18 & 0.4 \\
Return to library after first write (\%) & 22 & 30 & 47 \\
Agent steps & 15 & 27 & 47 \\
\bottomrule
\end{tabular}
\end{table}

\paragraph{Agents copy idioms rather than search for short names.} Only 4\% of trials grep for a prelude, \texttt{\_\_all\_\_}, or re-exports, and 62\% of greps for export declarations name a single symbol. Solutions still import from the library root or prelude in 91\% of Python, 70\% of Rust, and 47\% of Haskell cases, because they copy the documented idiom. Every \texttt{pandas} trial uses \texttt{import pandas as pd}, and \texttt{prelude::*} appears in 83\% of \texttt{rapier} and 93\% of \texttt{chumsky} trials. Deeper reading instead produces deeper imports. Agents import a symbol from where they found it (e.g., \texttt{bs4.element.Tag}), and deep imports where a short path exists rise from 4\% to 21\% of Rust trials between low and high effort.

\section{Explicit Guidance Prompt}\label{appendix:neuralese}
The prompt calls this agent-oriented design style \emph{neuralese}.

\paragraph{Reporting definitions.} \emph{Exported-name overlap} is the share of the production library's exported names that a generated library reproduces verbatim, averaged over a task's libraries and then over tasks. The \emph{simplicity and correctness contributions} in \autoref{fig:neuralese} split each paired cell's score change (the same implementer and problem under both prompts) into the part due to the change in simplicity and the part due to the change in test-pass fraction, using a symmetric two-factor split of the per-problem product in \autoref{eq:score}. Contributions are averaged over tasks and then over languages with equal weight. Under this split, the guidance prompt gains \neuralesesimplicitygain{} points from simplicity and loses \neuralesepassrateloss{} from correctness.

\begin{lstlisting}[style=lstprompt,caption={Explicit guidance prompt: the agent-oriented library-design condition of \autoref{sec:agent-preferences}.},label={lst:prompt-neuralese}]
{{instruction}}

## Who this library is for

Humans do not need to understand this library at all, only agents. Fresh coding agents. Each one gets a task in this domain, has `{{library_name}}` installed with its docs, and is scored on how few tokens of code it writes on top of the library while passing hidden tests. The library is a channel between you and that agent: you compress the domain into an API, the agent decompresses its task into a handful of calls. Every token the agent still has to write is a token your channel failed to carry.

Design in **neuralese**: the form two models would settle on if they only had to talk to each other. Optimise token economy for a model, not legibility for a person. Judge every design choice by one question: *what would an agent prefer?*

## What an agent prefers

- One verb per intent that carries the whole task: parse this document, resolve these references, render that report. A model states its intent in one line and wants one call that matches it.
- Names that are the intent, arguments that are the task's own nouns, results that are the shape the task asked for.
- Defaults that already match the common case, so the common program has no configuration at all; the uncommon case is one keyword away.
- Edge cases, validation, ordering, formatting and diagnostics inside the call. The agent writes the happy path and gets the correct program.
- Dense examples over prose. A model finds the example nearest its task and copies it; it reads reference docs only when no example fits.
- Big flat surfaces over layers. Human decomposition, small composable pieces, builders, class hierarchies, configuration objects and abstractions earn a place only where the agent's program gets shorter with them than without.

## Workflow

1. **Write the consumer's programs first.** From the example usages above and the tasks a library of this kind exists for, write ten to fifteen distinct downstream tasks as the program a model would most want to write, in `/workspace/SKETCHES.md`: three to eight lines each, calling functions that do not exist yet. Done when the set spans input parsing, the core operations, output shapes and error paths, and no sketch holds a loop, branch or helper the library could own.
2. **Design the API from the sketches.** Every function a sketch calls is public API with that name and signature. Done when each sketch type-checks against the design.
3. **Implement**, using only these dependencies: {{libraries}}.
4. **Ask the agents.** If you can spawn subagents, do it: for each sketch, hand a subagent only the downstream task text plus the library as installed and its README, with no memory of your design, and have it write the program. Where its program is longer than the sketch, guesses a name wrong, or has to read reference docs, fix the library, then ask again. If you cannot spawn subagents, do the same from a clean context yourself, task text and README only. Done when a fresh agent lands on the sketch without help, for every sketch.
5. **Freeze the examples.** Each sketch, unchanged, becomes a runnable file under `/workspace/examples/` and runs on realistic input. Done when every example runs.
6. **Write `/workspace/README.md`** for the consumer: the examples first, each with one line naming the task it solves, then the reference, then packaging.
7. **Package** as the task instructions specify, build offline, and submit.
\end{lstlisting}

\section{Timeouts}\label{appendix:timeouts}
Of the 28{,}314 \phasetwo{} trials (one implementer solving one problem with one library), 2.3\% ended through budget exhaustion or library-installation failure. The agent ran out of time or budget in 1.7\% of trials, and the authored library failed to install in 0.6\%.
We keep every such trial; when the agent runs out of time or budget, we grade whatever it left behind.
Dropping these trials instead raises the mean score by at most 1.2 points per arm and does not change the ordering of the no-library, agent-authored, and production-library arms.

\end{document}